\documentclass{article}

\PassOptionsToPackage{numbers, compress, square}{natbib}
     \usepackage[preprint]{neurips_2020}
\usepackage{graphicx}
\usepackage{subcaption}

\usepackage[utf8]{inputenc} 
\usepackage[T1]{fontenc}    
\usepackage{hyperref}       
\usepackage{url}            
\usepackage{booktabs}       
\usepackage{amsfonts}       
\usepackage{nicefrac}       
\usepackage{microtype}      
\usepackage{amsmath}

\title{Fractal Dimension Generalization Measure}

\author{%
  Valeri Alexiev\thanks{valeri.alexiev@gmail.com} \\
  CognitiveScale\\
  \texttt{valexiev@cognitivescale.com} \\
}

\begin{document}

\maketitle

\begin{abstract}
    Developing a robust generalization measure for the performance of machine learning models is an important and challenging task. A lot of recent research in the area focuses on the model decision boundary when predicting generalization. In this paper, as part of the "Predicting Generalization in Deep Learning" competition, we analyse the complexity of decision boundaries using the concept of fractal dimension and develop a generalization measure based on that technique. 
\end{abstract}

\section{Introduction}

Understanding and predicting the generalization capabilities of models is becoming more and more important as machine learning algorithms continue to proliferate in a variety of areas and impact an increasing number of people. The "Predicting Generalization in Deep Learning" competition (\citet{jiang2020competition}) presents a particular problem setting in which to compare the performance of different techniques for predicting the generalization ability of models. The goal is, given a model and its training data, to produce a measure of the generalization gap of the model. The competition focuses on convolutional neural networks for image classification tasks with the public dataset being composed of models trained on CIFAR-10 and SVHN. The metric used to rate the generalization prediction techniques is the conditional mutual information between the generalization measure and the true generalization gap conditioned on groups of hyperparameter settings (\citet{jiang2019fantastic}). 

Most generalization measures try to capture some concept of model or decision boundary complexity. Similarly, fractal dimension analysis has been applied in various areas of data science as a way to consider the complexity of datasets. In the present paper, we propose a generalization measure based on the concept of fractal dimension that compares the complexity of a model's decision boundary to the complexity of the training data. We discuss the behavior of the various hyperparameters of the technique. We also share results on two tasks from the competition and propose future directions of research. 

\section{Background}

\subsection{Generalization Measures}

The shape of the decision boundary is crucial to the generalization ability of a model. As such, the area of research focusing on decision boundaries is quite rich. \citet{yousefzadeh2019investigating} investigate how \textit{flip points}, points on the decision boundary, can provide information about the generalization error and robustness of a model. \citet{He2018DecisionBA} investigate how the nature of the decision boundary relates to adversarial examples and discover that the decision boundary of a classifier near adversarial examples is different than around regular examples. 

\citet{jiang2019predicting} propose a generalization measure that is based on the concept of margin distribution, which are the distances of training points to the decision boundary. However, a later survey by \citet{jiang2019fantastic} claims that sharpness-based measures, focusing on the concept of \textit{sharpness} of local minima, are the best performing family of generalization measures. 

\subsection{Fractal Dimension}

Fractal dimension is a measure of how an object scales differently than the space it’s embedded in and is a common way to analyze the complexity of objects. Despite the focus from the machine learning community on decision boundaries and their properties, not much research has been done on performing fractal dimension analysis on them. Fractal dimension analysis has been used for dataset analysis (\citet{bloem2010fractal}) and data mining (\citet{kumaraswamy2003fractal}), but no special focus has been placed on the fractal dimension of the decision boundaries of models themselves. In the present work, we're basing our generalization measure on the fractal dimension of decision boundaries as a way to measure the complexity of a model. 

\section{Methodology}

The proposed complexity measure is based on the concept of fractal dimension. The idea is to approximate the fractal dimension of the decision boundaries in the data as well as the decision boundaries of the provided model. The ratio between the two signifies how close the model is to capturing the structure of the dataset and, by extension, its generalization ability.
    
We approximate the fractal dimension through the concept of correlation dimension. We compute the correlation dimension of the class boundary by focusing on examples from different classes. For each batch of randomly-selected data samples, we compute the pairwise distances of samples in the batch and then count the number of pairs with different classes within distance $r$ for a range of distances $r \in R$. 
    \begin{equation}
        C(r) = \#(pairs\;within\;distance\;r)
    \end{equation}
The distances $r \in R$ are dataset-specific and are automatically generated from a range of percentiles, which are one of the hyperparameters of this technique. 
    
Once we have the counts for each distance $r$, we convert both the counts $C(r)$ and the distances $r$ to log-space and then fit a line to this data. The slope of the line is the correlation dimension of our data (\citet{kumaraswamy2003fractal,Attikos2009FasterEO}):
\begin{equation}
    D_2 = \frac{\delta \log(C(r))}{\delta \log(r)}, r \in R
\end{equation}
    
Because of the memory overhead of computing pairwise distances on the entire dataset, we calculate the fractal dimension based on pair counts within a single batch and then average the results of all batches to get our final estimate of the fractal dimension.
    
In order to explore the decision boundary of the model beyond the provided examples, we utilize the \textit{mixup} technique introduced by \citet{zhang2018mixup}. A data augmentation technique, \textit{mixup} is used to generate new virtual samples by linearly interpolating between pairs of real samples:
    \begin{equation}
    \begin{split}
        \tilde{x} = \lambda x_1 + (1 - \lambda)x_2\\
        \tilde{y} = \lambda y_1 + (1 - \lambda)y_2,
    \end{split}
    \end{equation}
where $(x_1, y_1)$ and $(x_2, y_2)$ are randomly-selected examples and $\lambda \in [0, 1]$. 

\begin{figure}[t]
  \centering
  \begin{subfigure}[t]{0.46\textwidth}
  \centering
  \includegraphics[trim=125 10 150 0,clip,width=8cm]{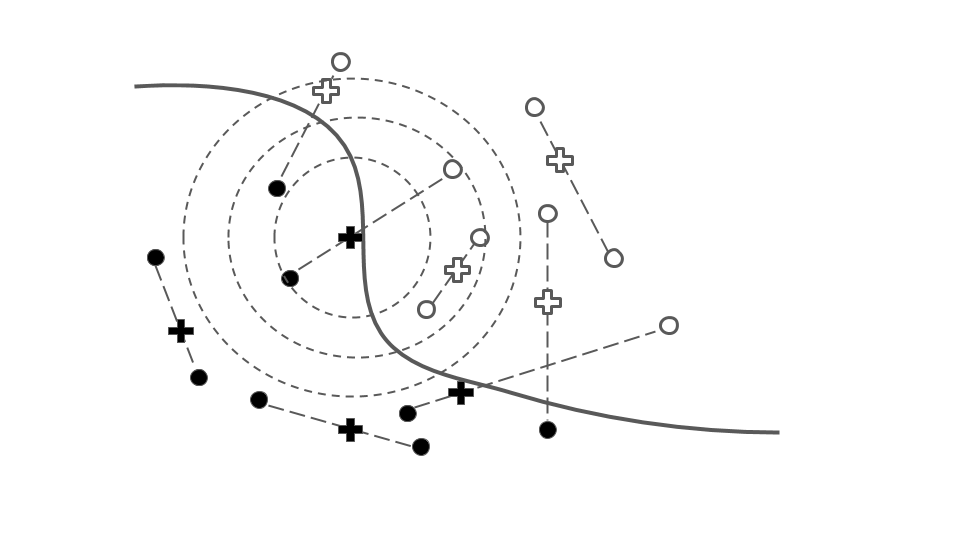}
  \caption{The white and black dots represent samples with different \textit{model outputs}. $+$s represent the virtual samples generated by the \textit{mixup} operation.}
  \end{subfigure}
  \quad
  \begin{subfigure}[t]{0.46\textwidth}
  \centering
  \includegraphics[trim=125 10 150 0,clip,width=8cm]{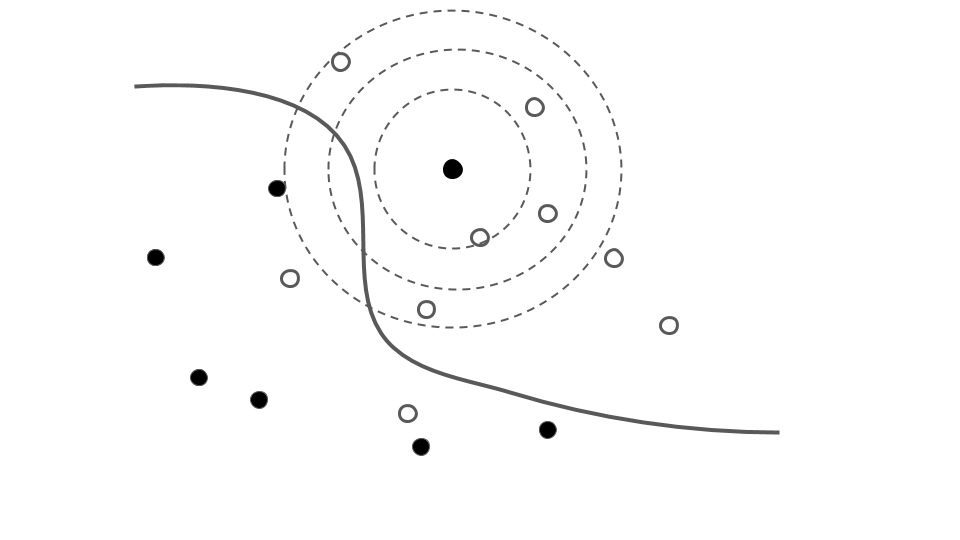}
  \caption{The white and black dots represent samples with different \textit{labels}.}
  \end{subfigure}
  \caption{Example of pair counting for $D_2(X_{mixup})$ (a) and $D_2(X)$ (b). The curve represents the model's decision boundary, while the concentric circles illustrate the range of $R$ used to count pairs for one particular sample.}
  \label{fig:pair-counting}
\end{figure}

Our proposed generalization measure $M$ is equal to the absolute difference between 1 and the ratio of the correlation dimension of the \textit{mixup}-augmented class boundary and the training data class boundary of the model under inspection:
    \begin{equation}
        M = \left\vert 1 - \frac{D_2(X_{mixup})}{D_2(X)} \right\vert,
    \end{equation}
where $D_2(X_{mixup})$ is the fractal dimension of the augmented data class boundary (using the model outputs as labels) and $D_2(X)$ is the fractal dimension of the training data class boundary (using the actual training data labels) (See Figure~\ref{fig:pair-counting}). The closer $M$ is to 0 the better the model decision boundary matches the complexity of the training data and, therefore, the better it generalizes. 

\section{Results}

\begin{table}
  \caption{Conditional Mutual Information scores on the public data}
  \label{results-table}
  \centering
  \begin{tabular}{llllll}
    \toprule
    Technique & Number of batches   & Batch size & R percentile range & Task 1 & Task 2 \\
    \midrule
    Jacobian Norm & - & - & - & 0.0094 & \textbf{0.031} \\
    Fractal Dimension & 100 & 128 & [0.01; 0.3] & \textbf{0.041} & 0.020    \\
    Fractal Dimension & 100 & 128 & [0.01; 0.5] & 0.026 & 0.018    \\
    Fractal Dimension & 150 & 128 & [0.01; 0.5] & 0.026 & 0.011    \\
    Fractal Dimension & 150 & 128 & [0.01; 0.3] & 0.033 & 0.020    \\
    Fractal Dimension & 100 & 96 & [0.01; 0.3] & 0.027 & 0.020    \\
    \bottomrule
  \end{tabular}
\end{table}

Table~\ref{results-table} shows the conditional mutual information score between $M$ and the actual test set performance, which is the baseline score designated by the competition (\citet{jiang2020competition}). A higher conditional mutual information means better ability of the technique to predict the generalization ability of the model. There are three hyperparameters in our computation of the fractal dimension generalization measure: batch size; the number of batches; and the $R$ percentile range which is the percentiles of inter-sample distances we use to calculate the set $R$ from the data.  

From our experiments, we conclude that the batch size and the R percentile range are both important parameters. Smaller batch sizes (such as 64) often fail to produce results as the number of samples is too small for the computation of the fractal dimension (those results are not included in Table~\ref{results-table}). We theorize that the minimal useful batch size is related to the dimensionality of the data, but further experimentation is necessary. Similarly, $R$ percentile ranges with higher upper bound tend to underperform. We theorize that this is because at larger scales the behavior of $D_2(X_{mixup})$ and $D_2(X)$ is always similar and, therefore, is not a good measure of the generalization ability of the model. Intuitively, we expected that an increase in the number of batches used for the computation of the fractal dimension would produce an improvement in the performance of our proposed measure. However, this is not the case. We suspect this indicates that the measure has large variance, but more experiments are needed to confirm this.

\section{Conclusion}

In this work, as part of the "Predicting Generalization in Deep Learning" competition, we proposed a generalization measure based on the concept of fractal dimension that works by comparing the complexity of the model decision boundary with the complexity of the data. While the approach is not competitive with the best performing techniques, it provides an alternative direction of research and has space for improvement in multiple areas. In particular, better ways to approximate the decision boundary of a model and the fractal dimension could improve the performance of the technique. In addition, more effort should be made into making the measure more stable as there is evidence that, in its current form, it has large variance.


\bibliography{sources}

\end{document}